\documentclass[times]{TRR}

\usepackage{moreverb,url}
\usepackage{enumitem}
\usepackage{amsmath,amssymb,amsfonts}
\usepackage{graphicx,color}
\usepackage{textcomp}
\usepackage{xcolor}
\usepackage{multirow}
\usepackage[table]{xcolor}
\usepackage{caption}
\usepackage{adjustbox}
\usepackage{algorithm}
\usepackage{algorithmic}
\usepackage{xcolor,colortbl}      
\usepackage[most]{tcolorbox}
\usepackage{tabularx}
\usepackage{newfloat}
\usepackage{listings}
\usepackage{soul}
\usepackage{color}
\usepackage{caption}
\usepackage{multirow}

\usepackage[colorlinks,bookmarksopen,bookmarksnumbered,citecolor=red,urlcolor=red]{hyperref}

\newcommand\BibTeX{{\rmfamily B\kern-.05em \textsc{i\kern-.025em b}\kern-.08em
T\kern-.1667em\lower.7ex\hbox{E}\kern-.125emX}}

\def\volumeyear{2020}

\begin{document}

\runninghead{Smith and Haynes}

\title{Retrieval-Augmented Framework for LLM-Based Clinical Decision Support}

\author{Leon Garza\affilnum{1}, Anantaa Kotal\affilnum{1}, Michael A. Grasso\affilnum{2}, Emre Umucu\affilnum{3}}

\affiliation{\affilnum{1}Dept. of Computer Science, The University of Texas at El Paso, USA\\
\affilnum{2} Dept. of Emergency Medicine, University of Maryland School of Medicine, USA\\
\affilnum{3} Dept. of Public Health Sciences, The University of Texas at El Paso, USA}

\corrauth{Anantaa Kotal, akotal@utep.edu}

\begin{abstract}
The increasing complexity of clinical decision-making, alongside the rapid expansion of electronic health records (EHR), presents both opportunities and challenges for delivering data-informed care. This paper proposes a clinical decision support system powered by Large Language Models (LLMs) to assist prescribing clinicians. The system generates therapeutic suggestions by analyzing historical EHR data, including patient demographics, presenting complaints, clinical symptoms, diagnostic information, and treatment histories. The framework integrates natural language processing with structured clinical inputs to produce contextually relevant recommendations. Rather than replacing clinician judgment, it is designed to augment decision-making by retrieving and synthesizing precedent cases with comparable characteristics, drawing on local datasets or federated sources where applicable. At its core, the system employs a retrieval-augmented generation (RAG) pipeline that harmonizes unstructured narratives and codified data to support LLM-based inference. We outline the system's technical components, including representation representation alignment and generation strategies. Preliminary evaluations, conducted with de-identified and synthetic clinical datasets, examine the clinical plausibility and consistency of the model's outputs. Early findings suggest that LLM-based tools may provide valuable decision support in prescribing workflows when appropriately constrained and rigorously validated. This work represents an initial step toward integration of generative AI into real-world clinical decision-making with an emphasis on transparency, safety, and alignment with established practices.
\end{abstract}

\maketitle

\section{Introduction}
Healthcare stands at a critical turning point.  The prevalence of chronic conditions is increasing, populations are aging, and diagnostic complexity continues to increase. At the same time, the volume of digital health data is expanding at an unprecedented pace. Among these resources, Electronic Health Records (EHRs) remain one of the most valuable yet underutilized tools in clinical care \cite{janssen2022electronic}.  These databases hold long-term in-depth data on patients, such as their demographics, presenting complaints, diagnostic tests, treatment plans, and results after follow-up.  However, they are often fragmented and predominantly unstructured, making real-time interpretation difficult, especially in settings with high patient volumes and limited clinical bandwidth \cite{li2022neural}.

Clinical prescribing extends beyond the adherence to treatment guidelines; it requires balancing broad medical knowledge with the individualized needs of each patient.  Under significant time constraints, clinicians must weigh co-occurring diseases, allergies, prior adverse reactions, and potential drug interactions.  This complexity, combined with cognitive workload and disjointed access to information, can lead to suboptimal prescribing decisions.  According to Gandhi et al. \cite{gandhi2003adverse} and Bates et al. \cite{bates1998effect}, prescribing errors remain a major cause of adverse drug events (ADEs), affecting up to 7\% of hospitalized patients and contributing to serious patient harm, prolonged hospitalizations, and increased healthcare costs.

To mitigate these risks, healthcare systems have adopted computerized physician order entry (CPOE) systems and clinical decision support systems (CDSSs) 
\cite{kaushal2003effects}. While these technologies reduce certain classes of medication errors, they are often built on rigid, rule-based logic that lacks adaptability and contextual sensitivity. Furthermore, most rely on structured input formats such as coded diagnoses or laboratory values, neglecting the rich clinical reasoning captured in free-text physician notes, discharge summaries, and consultation records. Consequently, clinicians often override alerts, disregard recommendations, or abandon these tools due to their nonspecificity, irrelevance, or usability challenges \cite{sutton2020overview}.

Advances in artificial intelligence (AI), particularly in LLMs, present promising alternatives \cite{lee2020clinical, shickel2017deep}.  LLMs like GPT-4 \cite{achiam2023gpt}, BioBERT \cite{lee2020biobert}, and Med-PaLM \cite{tu2024towards} have demonstrated strong performance in clinical summarization, question answering, information retrieval, and zero-shot reasoning \cite{singhal2022large}.  Trained on extensive biomedical datasets and fine-tuned on domain-specific datasets, LLMs are capable of interpreting complex clinical queries, integrating diverse information streams, and generating contextually appropriate outputs. Unlike traditional CDSSs, LLMs can process both structured and unstructured inputs, enabling them leverage the full breadth of EHR data.

Nevertheless, direct application of LLMs in clinical setting raises significant challenges. EHR data are heterogeneous, irregular, and often incomplete. Unprocessed clinical text can yield generic or irrelevant outputs if patient context is not adequately represented or grounded in  precedent. Moreover, the high-stakes nature of medical decision-making necessitates systems that are transparent, verifiable, and interpretable by human experts. These demands require architectures that go beyond end-to-end generation, incorporating mechanisms for retrieval,  structuring, and reasoning that support clinician support.

In this paper, we introduce a novel LLM-powered clinical decision support framework  designed to assist prescribers in generating safe and contextually appropriate treatment recommendations. Central to our system is a retrieval-augmented generation (RAG) pipeline with structured case comparison. The model constructs a composite patient profile, including demographics, presenting complaints, laboratory results, and narrative notes, and uses this profile to retrieve semantically and clinically similar historical cases from a local or federated database. These retrieved cases ground the LLM’s generative process, improving contextual relevance and clinical plausibility.

This approach addresses several goals. First, it ensures recommendations are not based solely on statistical priors learned during pretraining but are anchored in real, interpretable patient histories. Second, it enhances explainability by allowing clinicians to trace outputs back to precedent cases, thereby fostering trust and enabling oversight. Third, its modular and extensible design facilitates integration with existing EHR platforms and adaptation across specialties and institutional workflows.

The system is intended to support prescribing across diverse contexts, from primary care to specialist decision-making. It is particularly valuable in cases involving diagnostic uncertainty, polypharmacy, or rare conditions, settings where precedent and comprehensive patient modeling can reduce ambiguity. In high-volume environments such as emergency departments or telehealth platforms, the framework can assist with initial triage or flag outlier cases requiring additional review. This work is guided by the following research questions:

\begin{itemize}[leftmargin=*]
    \item \textbf{RQ1:} How can large language models be effectively conditioned on heterogeneous EHR inputs—both structured and unstructured—to support prescribing decisions in real-world clinical settings?
    \item \textbf{RQ2:} What retrieval strategies are most effective for identifying clinically analogous cases from historical data to inform the model’s recommendations?
    \item \textbf{RQ3:} How can we evaluate the clinical plausibility, safety, and reliability of LLM-generated prescribing suggestions in a rigorous and domain-sensitive manner?
\end{itemize}

The proposed framework is designed to complement—not replace—clinical expertise. By leveraging prior patient outcomes and treatment pathways, it provides a supportive analytic layer that enhances decision-making in high-uncertainty or high-risk contexts. This aligns with the paradigm of ``human-in-the-loop'' AI, where clinicians remain central decision-makers, augmented by data-driven insights.

The \textbf{key contributions} of this paper are as follows:

\begin{itemize}
    \item \textbf{Design of an LLM-based prescribing support tool} that unifies structured and unstructured EHR data into a composite patient representation.
    \item \textbf{Implementation of a RAG architecture} that grounds outputs in semantically and clinically similar prior cases, improving factual grounding and relevance.
    \item \textbf{Development of a preprocessing and alignment pipeline} that transforms raw EHR data—including clinical notes, laboratory values, ICD codes, and medication history—into embeddings suitable for LLM inference.
    \item \textbf{Discussion of implementation challenges and deployment considerations}, including transparency, explainability, auditability, and compliance with regulatory standards for clinical AI.
\end{itemize}

By addressing these challenges, this work advances research at the intersection of clinical informatics and machine learning. It emphasizes that effective decision support systems must not only be accurate but also contextually aware, explainable, and seamlessly integrated into clinical workflows. Unlike prior approaches that oversimplify patient data or disregard unstructured content, our framework adopts a holistic perspective, using LLM reasoning to generate patient-specific insights that are both data-driven and human-centered.

Ultimately, we aim to demonstrate the feasibility and value of LLM-based decision support for safer, more consistent, and more personalized prescribing. As these technologies mature, their responsible integration into clinical practice can serve as a catalyst for more equitable, personalized, and effective healthcare delivery.

\section{Methodology}
\subsection{Task Definition}

The core objective of this work is to support clinical prescribers by generating context-aware treatment recommendations based on patient-specific data derived from EHRs. We formalize this as a retrieval-augmented conditional generation task, wherein a language model generates therapeutic suggestions grounded in prior, semantically and clinically analogous cases.

Let a patient record $P$ be composed of structured features $P_s$ and unstructured features $P_u$. The structured features include tabular EHR elements such as demographics (age, sex, race), diagnostic codes (e.g., ICD-10), laboratory values, vital signs, allergies, and medication history. The unstructured features include clinical narratives such as physician notes, history of present illness (HPI), discharge summaries, and assessment plans. Together, $P = \{P_s, P_u\}$ represents the full patient context.

The task is to generate a ranked list of plausible treatment recommendations $\hat{T} = \{t_1, t_2, \dots, t_n\}$, conditioned on the patient record $P$ and additional retrieved evidence $C$ from a case corpus $\mathcal{D}$:
\[
\hat{T} = \mathcal{M}(P, C), \quad \text{where } C = \text{Retrieve}(P, \mathcal{D})
\]
Here, $\mathcal{M}$ denotes the generative language model (e.g., T5, GPT), and $\text{Retrieve}$ is a similarity-based function that returns $k$ most relevant historical cases based on patient similarity. The retrieved set $C$ provides precedent-based grounding to the generative model, enhancing the clinical plausibility and traceability of outputs.

This task differs from traditional clinical question answering (QA) or diagnosis prediction in several ways:
\begin{itemize} [leftmargin=*]
    \item \textbf{Prescribing is action-oriented:} The goal is not to identify a condition but to recommend a safe, effective treatment plan.
    \item \textbf{Precedent grounding:} Recommendations are expected to reflect institutional knowledge or historical cases, not generic best practices.
    \item \textbf{Input heterogeneity:} The system must jointly process structured codes and free-form text, each containing critical cues for decision-making.
    \item \textbf{Safety-critical constraints:} Outputs must be clinically plausible and aligned with accepted standards, tolerating neither hallucination nor omission of contraindications.
\end{itemize}

We emphasize that the system is designed as a decision support tool rather than a prescriptive authority. The generated suggestions are meant to inform—not override—clinical judgment, with transparency and traceability as first-class principles.

\begin{figure*}[ht]
    \centering
    \includegraphics[width=\linewidth]{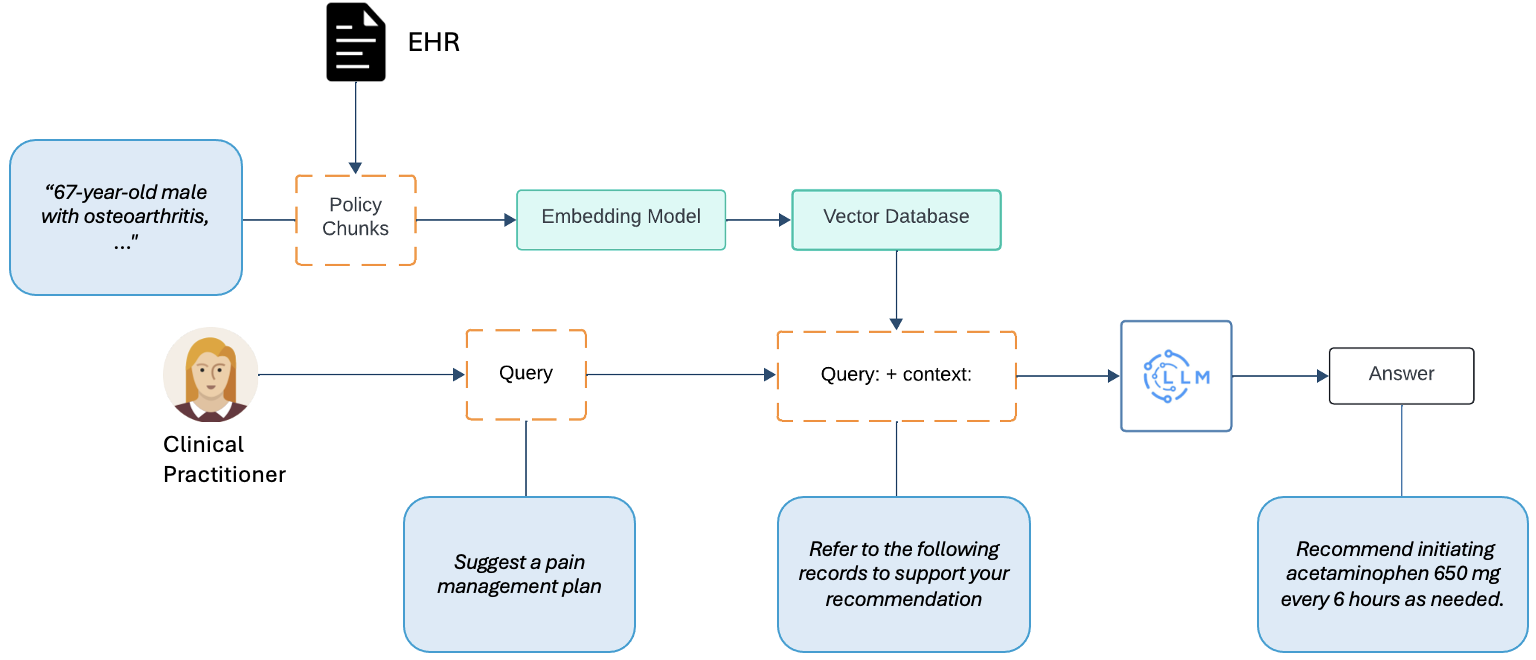}
    \caption{Overview of the proposed prescribing support architecture. Structured and unstructured EHR data are preprocessed and encoded into embeddings. Relevant historical cases are retrieved using a similarity search, then combined with the current patient profile to form an augmented prompt for the language model. The LLM generates a ranked list of recommended treatments, optionally flagged for safety checks.}
    \label{fig:system_arch}
\end{figure*}

\subsection{System Architecture Overview}

The proposed system is a modular pipeline that integrates structured and unstructured EHR data using a RAG framework to assist in clinical prescribing. The architecture consists of five core components that work together to transform raw patient data into context-aware, precedent-grounded treatment recommendations.
\begin{itemize} [leftmargin=*]
    \item \textbf{Data Ingestion and Preprocessing:} The system accepts structured (e.g., demographics, lab values, medication history) and unstructured (e.g., clinical notes, discharge summaries) EHR inputs. Data is normalized, temporally ordered, and tokenized. Unstructured text is segmented into clinically meaningful sections such as history of present illness (HPI), assessment, and plan. This preprocessing ensures consistent representation across patient records.

    \item \textbf{Patient Representation:} Structured and unstructured features are fused into a unified representation. Domain-adapted embedding models (e.g., BioBERT, Clinical SBERT) encode the patient context into a dense vector embedding. This embedding serves as the input query for retrieval and generation.

    \item \textbf{Case Retrieval:} A similarity-based retriever searches a vector index of previously encoded patient cases to find the top-$k$ most similar records. Each retrieved case includes both its context and the prescribed treatments. Retrieval is based on cosine similarity in the embedding space and is designed to surface semantically and clinically relevant precedents.

    \item \textbf{Prompt Construction:} Retrieved cases are combined with the current patient’s data to form a structured prompt. Prompts are assembled using standardized templates with special tokens to separate query, context, and evidence. Care is taken to ensure the prompt fits within the LLM’s token limit while retaining relevant clinical information.

    \item \textbf{Language Model Generation:} A pretrained or instruction-tuned LLM (e.g., T5, LLaMA2) processes the constructed prompt and generates a ranked list of recommended treatments. Outputs may include rationales, references to retrieved cases, and optional confidence indicators. This stage completes the RAG loop by synthesizing context, precedent, and clinical reasoning into actionable suggestions.
\end{itemize}

The system’s modular design supports flexible substitution of components. For instance, the retriever can be upgraded independently of the generator, or a different embedding model can be used depending on the clinical domain. This design ensures adaptability to evolving clinical needs, institutional practices, and model advancements.

\subsection{Data Ingestion and Preprocessing}

To enable effective retrieval and generation, patient data from EHRs must first be ingested, normalized, and converted into a representation suitable for semantic search and prompt construction. Our system handles both structured and unstructured EHR inputs through a multi-stage preprocessing pipeline.

\noindent \textbf{Structured Data Normalization:} Structured elements such as demographics (age, sex, race), vital signs, laboratory test results, problem lists, allergies, and medication history are extracted from standard EHR schemas. Terminologies are normalized using common clinical vocabularies: ICD-10 for diagnoses, LOINC for lab results, and RxNorm for medications. Temporal metadata (e.g., encounter timestamps, lab result times) are preserved to retain clinical sequence.

\noindent  \textbf{Unstructured Data Segmentation:} Clinical notes are parsed into semantically meaningful sections, such as “Chief Complaint,” “History of Present Illness,” “Assessment,” and “Plan,” using regular expression-based heuristics and section-header classifiers. Sentence-level tokenization is applied to support downstream chunking and embedding. Abbreviations and shorthand terms are optionally expanded using medical dictionaries to improve interpretability for embedding models.

\noindent \textbf{Data De-identification:} All patient identifiers (names, locations, dates, medical record numbers) are redacted or replaced using deterministic anonymization routines. In scenarios involving real patient data, this step ensures compliance with privacy regulations such as HIPAA. For synthetic datasets, identifiers are consistently randomized.

\noindent \textbf{Temporal Ordering and Windowing:} Clinical events are organized in temporal order, and time windows are optionally applied to restrict context to a recent or relevant span (e.g., past 90 days). This is critical for modeling acute vs. chronic conditions and ensuring that retrieval and generation are temporally aligned with the clinical state.

\noindent \textbf{Data Chunking:} Both structured and unstructured content are segmented into manageable, semantically coherent chunks for embedding. For example, a progress note might be divided into the assessment and plan sections, while lab results may be grouped into logical panels (e.g., metabolic panel, complete blood count). Each chunk is tagged with its source type and timestamp.

\noindent  \textbf{Embedding Preparation:} All preprocessed chunks are passed through an encoder (e.g., BioBERT, Clinical SBERT) to generate dense vector embeddings. These embeddings are stored in a vector database and serve as inputs to the retrieval module. Metadata is preserved alongside embeddings to support interpretability and filtering during retrieval.

This preprocessing pipeline ensures that both structured and unstructured clinical signals are harmonized into a consistent format suitable for RAG. It also supports future extensions to handle multimodal data, such as imaging or genomics, by incorporating new embedding strategies and chunking logic.

\subsection{Retrieval-Augmented Generation Framework}

LLMs possess powerful generative capabilities, but when applied to clinical domains such as prescribing, they often suffer from factual hallucination, lack of specificity, and contextual drift. To address these limitations, we adopt a RAG framework, which combines language generation with a retrieval mechanism grounded in historical EHR data. This hybrid architecture enables the model to generate treatment suggestions that are both fluent and anchored in precedent.

Unlike traditional LLMs that rely solely on pre-trained knowledge, RAG models dynamically incorporate external evidence at inference time. In our setting, this evidence consists of prior patient cases that are semantically and clinically similar to the current input. The retrieved cases serve two purposes: they constrain the generative space to plausible decisions and provide justifiable precedents for each suggestion.

\subsubsection*{Retriever Component: Identifying Relevant Clinical Cases}

The retriever is tasked with identifying the top-$k$ most relevant cases from a pre-indexed corpus $\mathcal{D}$ of historical patient records. Each case is preprocessed into semantically meaningful chunks and embedded using a clinical encoder such as BioBERT or Clinical SBERT. These embeddings are stored in a vector database (e.g., FAISS), enabling fast similarity-based retrieval.

\begin{itemize}
    \item \textbf{Query Embedding:} The current patient input $P = \{P_s, P_u\}$ is embedded into a dense vector $E_P$ using the same encoder as the corpus.
    \item \textbf{Vector Search:} Cosine similarity is computed between $E_P$ and all candidate embeddings $\{E_{d_i}\}_{i=1}^{|\mathcal{D}|}$. The top-$k$ most similar cases are selected as the retrieval set $C = \{d_1, d_2, \dots, d_k\}$.
    \item \textbf{Semantic Filtering:} Optionally, retrieved cases can be filtered or reranked based on constraints such as diagnosis overlap, temporal recency, or medication class.
\end{itemize}

Each retrieved case includes both its input context and prescribed treatments, enabling the downstream model to associate input patterns with clinical actions.

\subsubsection*{Generator Component: Producing Contextualized Recommendations}

The generator model $\mathcal{M}$ is a pretrained LLM (e.g., T5, LLaMA2) conditioned on a structured prompt consisting of:
\begin{itemize}
    \item The current patient summary (structured + unstructured).
    \item The $k$ retrieved cases with their associated treatments.
    \item An instruction template specifying the generation task (e.g., ``Suggest a treatment plan for the following patient based on similar historical cases.'').
\end{itemize}

The prompt is formatted using special tokens to clearly delineate between the patient query and retrieved evidence. An example template is shown below:

\begin{tcolorbox}[enhanced,
  attach boxed title to top center={yshift=-3mm, yshifttext=-1mm},
  colback=white,
  colframe=gray!75!black,
  colbacktitle=gray!80!black,
  title=Clinical Prompt Example,
  boxed title style={size=small, colframe=gray!90!black},
  left=1mm,
  right=1mm,
  boxrule=0.75pt
]
\label{box:clinical_prompt}
\small

\textbf{Instruction:}\\
\textcolor{teal}{%
  Suggest a pain management plan for the following patient based on similar historical cases.
}

\vspace{3mm}
\textbf{Query Input:}\\
\textcolor{red!45!black}{%
  65-year-old male with osteoarthritis and chronic lower back pain, reporting pain severity 7/10 and limited mobility. No history of GI bleeding or opioid use.
}

\vspace{3mm}
\textbf{Retrieved Context:}\\
\textcolor{blue!65!black}{%
  67-year-old male with osteoarthritis and knee pain, reports pain severity 6/10, no prior GI issues or opioid exposure. Recommend initiating acetaminophen 650 mg every 6 hours as needed. \\
  63-year-old male with chronic lumbar pain, pain score 8/10, no GI history, not currently on analgesics. Recommend initiating acetaminophen 650 mg every 6 hours as needed. 
}

\vspace{3mm}
\textbf{Generated Output:}\\
\textcolor{green!45!black}{%
  Recommend initiating acetaminophen 650 mg every 6 hours as needed. 
}

\end{tcolorbox}

The model processes the prompt using self-attention mechanisms that relate symptoms, labs, and history in the query to the corresponding features in the retrieved cases. This architecture enables the model to generalize across patient variations while remaining anchored in empirical precedent.

\subsubsection{RAG Inference Algorithm}

The complete workflow for RAG inference is summarized in Algorithm~\ref{algo:rag}. Each step is modular, allowing substitution of the encoder, vector store, or generator.

\begin{algorithm}
\caption{RAG for Clinical Prescribing}
\label{algo:rag}
\begin{algorithmic}
\STATE \textbf{Input:} Patient Record $P$, Case Corpus $\mathcal{D}$, Embedding Model $\mathcal{E}$, Generator Model $\mathcal{M}$, Similarity Threshold $\tau$
\STATE \textbf{Output:} Treatment Recommendation $\hat{T}$, Supporting Cases $\mathcal{C}$

\STATE \textbf{Step 1: Patient Embedding}
\STATE $E_P \gets \mathcal{E}(P)$ \hfill \textit{// Compute embedding for patient record $P$}

\STATE \textbf{Step 2: Case Retrieval}
\STATE $\mathcal{C} \gets \emptyset$ \hfill \textit{// Initialize retrieved case set}
\FOR{each case $d_i \in \mathcal{D}$}
    \STATE $E_{d_i} \gets \mathcal{E}(d_i)$ \hfill \textit{// Compute embedding for historical case $d_i$}
    \STATE $sim \gets \text{cosine}(E_P, E_{d_i})$ \hfill \textit{// Compute similarity}
    \IF{$sim \geq \tau$}
        \STATE $\mathcal{C} \gets \mathcal{C} \cup \{d_i\}$ \hfill \textit{// Add case to retrieved set}
    \ENDIF
\ENDFOR

\STATE \textbf{Step 3: Prompt Construction}
\STATE $S \gets P \cup \mathcal{C}$ \hfill \textit{// Concatenate patient record with retrieved cases}

\STATE \textbf{Step 4: Recommendation Generation}
\STATE $\hat{T} \gets \mathcal{M}(S)$ \hfill \textit{// Generate treatment recommendation}

\RETURN $\hat{T}, \mathcal{C}$
\end{algorithmic}
\end{algorithm}

In our setup, to support traceability and accountability, we consider the following design considerations for our RAG based LLM QA architecture:
\begin{itemize}[leftmargin=*]
    \item \textbf{Context Window Limitations:} For long EHR narratives or numerous retrieved cases, prompt length may exceed the LLM’s context window. We mitigate this using summarization heuristics and rank-based truncation.
    \item \textbf{Case Diversity:} Retrieved cases must balance semantic similarity with diversity to avoid echoing a single treatment pattern.
    \item \textbf{Traceability:} Outputs can optionally cite the specific case (e.g., Case 2) that supports each recommendation, aiding interpretability.
\end{itemize}

This RAG framework provides a scalable and explainable foundation for leveraging LLMs in clinical prescribing. It enables the system to make context-aware, historically grounded recommendations that reflect the complex interplay of symptoms, demographics, and treatment trajectories common in real-world healthcare.

\section{Evaluation}
\subsection{Dataset}
We evaluate our clinical prescribing framework using a real-world dataset of emergency department (ED) encounters, previously collected and analyzed in a study on opioid and antimicrobial prescribing among insured and uninsured patients \cite{grasso2024opioid}. The dataset comprises 68,969 ED visits recorded over a two-year period (January 2017 to December 2018) at the University of Maryland Medical Center (UMMC), an academic tertiary care hospital in Baltimore, Maryland. \textbf{Regulatory approval} for this work was obtained from the University of Maryland School of Medicine Institutional Review Board.

Each ED encounter includes structured demographic and clinical attributes such as age, gender, race, housing status (e.g., homeless vs. housed), insurance status (insured vs. uninsured), comorbidity count, emergency severity index (ESI), and recidivism (total ED visits during the study period). In addition to structured data, each record contains unstructured fields such as chief complaints, provider notes, and discharge summaries, which are used to contextualize clinical decisions within our retrieval-augmented generation (RAG) architecture.

The primary prediction tasks are formulated as binary classification problems over three clinically meaningful outcome labels:
\begin{itemize}
\item \textbf{Recommended Non-Opioid Painkiller:} Indicates whether a non-opioid analgesic (e.g., ibuprofen, acetaminophen, naproxen) was prescribed during the ED visit.
\item \textbf{Recommended Opioid Painkiller:} Captures whether any opioid medication (e.g., oxycodone, hydrocodone, tramadol) was prescribed at discharge, regardless of dosage.
\item \textbf{Recommended Opioid Painkiller at Standard Dosage:} A subset of the above label, this indicates whether an opioid prescription was issued that adheres to safety guidelines (e.g., no more than a 3-day supply at the lowest dosage).

\end{itemize}

For preprocessing, structured features are normalized through binning or one-hot encoding where appropriate. Unstructured clinical texts are segmented into semantically coherent chunks (100–200 tokens) and embedded using domain-specific encoders. All records are indexed into a FAISS vector store to support similarity-based retrieval. The dataset is stratified and split into 70\% training, 15\% validation, and 15\% testing sets. Splits are designed to preserve the distribution of insurance status and prescription labels across folds while avoiding patient overlap.

\subsection{Baseline Comparisons}
To isolate the gains from unstructured text understanding and retrieval-based augmentation, we compare our system with classical machine learning baselines trained solely on structured patient features (e.g., demographics, acuity, comorbidities). The baseline models include:

\begin{itemize}
    \item Logistic Regression (LR)
    \item Decision Tree (DT)
    \item Random Forest (RF)
    \item Gradient Boosted Classifier (GBC)
    \item Support Vector Machine (SVM)
\end{itemize}

All models are trained and evaluated on identical data splits to ensure fair comparison. This setup quantifies the added value of LLM-based, case-grounded prescription reasoning.

\begin{table*}[]
\centering
\caption{Model performance on ED prescribing tasks (best results in \textcolor{blue}{blue})}
\label{tab:performance}
\resizebox{0.85\textwidth}{!}{
\begin{tabular}{p{4cm}lccccc}
\toprule
\textbf{Task} & \textbf{Model} & \textbf{Accuracy} & \textbf{Precision} & \textbf{Recall} & \textbf{F1} & \textbf{AUROC} \\
\midrule

\multirow{6}{=}{\raggedright\arraybackslash Recommended Opiod Painkiller} 
& RAG-LLM & 0.86 & 0.85 & 0.83 & 0.84 & \textcolor{blue}{0.91} \\
& Logistic Regression & 0.80 & 0.79 & 0.78 & 0.78 & 0.86 \\
& Decision Tree & 0.78 & 0.76 & 0.77 & 0.76 & 0.84 \\
& Random Forest & 0.79 & 0.78 & 0.78 & 0.78 & 0.85 \\
& Gradient Boost & \textcolor{blue}{0.87} & \textcolor{blue}{0.86} & \textcolor{blue}{0.84} & \textcolor{blue}{0.85} & 0.90 \\
& SVM & 0.81 & 0.80 & 0.81 & 0.80 & 0.89 \\
\midrule

\multirow{6}{=}{\raggedright\arraybackslash Recommended Opiod Painkiller at Standard dosage} 
& RAG-LLM & \textcolor{blue}{0.90} & \textcolor{blue}{0.91} & \textcolor{blue}{0.88} & \textcolor{blue}{0.89} & \textcolor{blue}{0.93} \\
& Logistic Regression & 0.88 & 0.87 & 0.88 & 0.87 & 0.90 \\
& Decision Tree & 0.87 & 0.86 & 0.87 & 0.86 & 0.89 \\
& Random Forest & 0.87 & 0.86 & 0.87 & 0.86 & 0.89 \\
& Gradient Boost & 0.88 & 0.88 & 0.88 & 0.88 & 0.91 \\
& SVM & 0.88 & 0.88 & 0.88 & 0.88 & 0.91 \\
\midrule

\multirow{6}{=}{\raggedright\arraybackslash Recommended Non-Opiod Painkiller} 
& RAG-LLM & 0.84 & \textcolor{blue}{0.86} & \textcolor{blue}{0.83} & \textcolor{blue}{0.84} & \textcolor{blue}{0.90} \\
& Logistic Regression & 0.83 & 0.82 & 0.81 & 0.81 & 0.87 \\
& Decision Tree & 0.81 & 0.80 & 0.80 & 0.80 & 0.86 \\
& Random Forest & 0.81 & 0.80 & 0.79 & 0.80 & 0.86 \\
& Gradient Boost & \textcolor{blue}{0.85} & 0.84 & 0.82 & 0.83 & 0.89 \\
& SVM & 0.82 & 0.81 & 0.81 & 0.81 & 0.87 \\

\bottomrule
\end{tabular}
}
\end{table*}

\subsection{Evaluation Metrics}
\subsubsection{Predictive Performance on Binary Classification Tasks}
To evaluate the effectiveness of the prescribing recommendations generated by our system, we frame each task (e.g., predicting opioid or non-opioid prescriptions) as a binary classification problem. We report the following standard metrics:

\begin{itemize}
    \item \textbf{Accuracy:} Overall proportion of correct predictions across the test set.
    \item \textbf{Precision:} Fraction of relevant instances among the retrieved instances; i.e., the proportion of true positives among predicted positives.
    \item \textbf{Recall (Sensitivity):} Fraction of relevant instances that were retrieved; i.e., the proportion of true positives among actual positives.
    \item \textbf{F1 Score:} Harmonic mean of precision and recall, balancing both false positives and false negatives.
    \item \textbf{AUROC (Area Under the ROC Curve):} Measures the model’s ability to distinguish between classes across all thresholds, especially valuable in settings with class imbalance.
\end{itemize}

Table~\ref{tab:performance} summarizes the performance of all evaluated models across three clinically relevant prediction tasks. We report Accuracy, Precision, Recall, F1 score, and AUROC. Best values for each metric are highlighted in \textcolor{blue}{blue}.

\subsubsection{Clinical Consistency of Recommendations}

To evaluate the clinical reliability of generated treatment plans, we adopt a \textit{relaxed agreement protocol} that assesses alignment with empirical prescribing behavior. A generated recommendation $\hat{T}$ for a patient input $P$ is considered \textit{clinically consistent} if it satisfies either of the following criteria:

\begin{itemize}
    \item \textbf{Exact Match:} The predicted treatment matches the actual prescription recorded in the EHR, i.e., $\hat{T} = T^*$.
    \item \textbf{Justified Deviation:} The predicted treatment $\hat{T}$ is not identical to $T^*$ but is supported by a retrieved historical case $d_i \in \mathcal{C}$ such that:
    \begin{enumerate}
        \item $\hat{T} \in d_i.T$ where $d_i.T$ is the treatment set in $d_i$, and
        \item $\text{sim}(P, d_i) \geq \tau$ based on a semantic similarity score computed over demographics, presenting complaint, and diagnosis.
    \end{enumerate}
\end{itemize}

To operationalize this:
\begin{enumerate}
    \item We encode both the patient query $P$ and each retrieved case $d_i \in \mathcal{C}$ using Clinical SBERT to compute cosine similarity.
    \item We set $\tau = 0.80$ as a similarity threshold based on domain validation.
    \item Each case that meets the above conditions contributes to the final Clinical Consistency Rate (CCR).
\end{enumerate}

We report two metrics: (i) the \textbf{CCR}, which includes both exact matches and justified deviations, and (ii) the \textbf{Exact Match Rate}, where only $\hat{T} = T^*$ counts as valid. The results are provided in Table \ref{tab:ccr}.

\begin{table}[h]
\centering
\caption{Clinical Consistency Rate under Relaxed Agreement Protocol}
\label{tab:ccr}
\resizebox{\columnwidth}{!}{
\begin{tabular}{lcc}
\toprule
\textbf{Model} & \textbf{CCR (\%)} & \textbf{Exact Match Only (\%)} \\
\midrule
Logistic Regression & 68.0 & 52.0 \\
Decision Tree       & 65.0 & 48.0 \\
Random Forest       & 70.0 & 55.0 \\
Gradient Boost      & 78.0 & 59.0 \\
SVM                 & 72.0 & 57.0 \\
RAG-LLM             & \textcolor{blue}{\textbf{82.0}} & \textcolor{blue}{\textbf{61.0}} \\
\bottomrule
\end{tabular}
}
\end{table}

\subsubsection*{Retrieval Module Effectiveness}
Given the dependency of our prescribing assistant on a RAG framework, the performance of the retrieval component is critical. To assess this, we evaluate the retriever's ability to surface clinically relevant analogues using two metrics:

\begin{itemize}
    \item \textbf{Top-$k$ Precision:} For each query patient record $P$, we measure the proportion of retrieved cases $\{d_1, d_2, \dots, d_k\}$ that share the same class label (i.e., treatment class) as the ground-truth prescription $T^*$. Formally:
    \[
        \text{Precision@}k = \frac{1}{k} \sum_{i=1}^k \mathbb{1}[d_i.T = T^*]
    \]
    \item \textbf{Mean Embedding Similarity:} We compute the average cosine similarity between the patient query embedding $E_P$ and each retrieved case embedding $E_{d_i}$:
    \[
        \text{MeanSim@}k = \frac{1}{k} \sum_{i=1}^{k} \cos(E_P, E_{d_i})
    \]
    where embeddings are obtained using Clinical SBERT trained on clinical narratives.
\end{itemize}

We evaluate these metrics for $k = \{3, 5, 10\}$ across all test queries in the corpus. Higher precision indicates retrieval of therapeutically aligned cases, while higher similarity reflects semantic coherence in the latent representation space.

\begin{table}[h]
\centering
\caption{Retrieval Quality Metrics at Varying Top-$k$ Levels}
\label{tab:retrieval}
\resizebox{\columnwidth}{!}{
\begin{tabular}{lccc}
\toprule
\textbf{Metric} & \textbf{Top-3} & \textbf{Top-5} & \textbf{Top-10} \\
\midrule
Precision@k (\%)     & \textbf{71.2} & 68.4 & 63.1 \\
MeanSim@k (cosine)   & 0.831 & \textbf{0.844} & 0.836 \\
\bottomrule
\end{tabular}}
\end{table}

\subsection*{Summary of Results}
Our results reveals that the proposed RAG-LLM framework delivers consistently strong performance across multiple prescribing tasks when compared to conventional machine learning models. For opioid prescribing at standard dosage, RAG-LLM achieved the highest accuracy (0.90), F1 score (0.89), and AUROC (0.93), indicating its ability to generate clinically appropriate and safety-aligned recommendations. In the broader opioid prescribing tasks, Gradient Boost slightly outperformed RAG-LLM on accuracy (0.87 vs. 0.86) and F1 score (0.85 vs. 0.84), while RAG-LLM retained the highest AUROC (0.91), suggesting superior robustness across thresholds. For non-opioid painkiller recommendations, RAG-LLM again led on F1 (0.84) and AUROC (0.90), with Gradient Boost achieving marginally higher accuracy (0.85).

Clinical reliability was further assessed using a relaxed agreement protocol. RAG-LLM achieved the highest Clinical Consistency Rate (CCR) at 82\%, with 61\% exact match to recorded prescriptions. These results indicate that the framework not only reproduces real-world prescribing patterns but also generates medically justifiable alternatives supported by retrieved historical cases.

Evaluation of retrieval quality confirmed the semantic and therapeutic relevance of the retrieval pipeline. RAG-LLM achieved a Precision@3 of 71.2\% and maintained a Mean Embedding Similarity above 0.83 across all top-k values.  Together, these findinga affirm the potential of retrieval-augmented generation for to support precise, transparent, and data-aligned clinical prescribing.

\section{Related Work}

The application of language models in clinical decision support lies at the intersection of machine learning, biomedical informatics, and healthcare delivery. In this section, we review relevant literature across four domains: (i) automated clinical decision support systems, (ii) language models in healthcare NLP, (iii) adaptation and training strategies for LLMs, and (iv) architectural innovations relevant to clinical use cases.

\subsection{Automated Clinical Decision Support Systems}
Traditional clinical decision support systems (CDSSs) supported clinician decision-making through rule-based alerts, reminders, and guideline-driven recommendations. These systems relied on manually encoded knowledge or decision trees derived from expert consensus and statistical heuristics. Examples include tools for detecting drug-drug interactions or contraindications using structured medication data \cite{sutton2020overview}. While effective in narrow context, such systems suffer from high false-positive rates, alert fatigue, and limited ability to handle complex or ambiguous inputs \cite{wright2018clinical}.

Subsequent approaches applied classical machine learning methods—such as logistic regression and gradient-boosted trees—to predict patient risks and treatment pathways from structured EHR data \cite{rajkomar2018scalable}. These models improved predictive accuracy but remain constrained by rigid tabular formats and limited interpretability.

With advances in deep learning and neural NLP, unstructured data such as clinical notes, radiology reports, and discharge summaries have been increasingly leveraged. Surveys by Shickel et al. \cite{shickel2017deep} and Li et al. \cite{li2022neural} highlight the use of recurrent architectures, transformers, and attention mechanisms to capture temporal dynamics and clinical context, improving performance in tasks like phenotyping, risk stratification, and early warning detection.

\subsection{Language Models in Clinical NLP}
The pretrained language models have reshaped biomedical NLP. Early domain-specific models such as BioBERT \cite{lee2020biobert}, ClinicalBERT \cite{huang2019clinicalbert}, and BlueBERT \cite{peng2020empirical} adapted the BERT architecture to biomedical and clinical corpora, achieving strong results on tasks such as named entity recognition, relation extraction, and document classification.

Generative LLMs have further extended these capabilities. Med-PaLM \cite{tu2024towards} and Med-PaLM 2 \cite{singhal2022large} have adapted instruction-tuned LLMs for open-ended clinical QA and reasoning, reaching expert-level performance on standardized exams. Similarly, GatorTron \cite{yang2022gatortron} scaled healthcare-specific LLMs to hundreds of billions of parameters. However, most such systems are evaluated on curated benchmarks rather than real-world treatment recommendation tasks. 

Recent work explores embedding LLMs into clinical workflows. Recent work by Agrawal et al. \cite{agarwal2024medhalu} explored LLMs for diagnostic reasoning in primary care, while others focused on clinical documentation, summarization \cite{singhal2025toward}, or medication reconciliation \cite{nori2023capabilities, gilson2023does}. The Clinical Camel project fine-tuned a multi-modal LLM for diagnostic suggestion using case vignettes \cite{toma2023clinical}, though without frounding in real-world EHRs or prescribing pathways. Despite progress, integration of LLM- and NLP-based tools based CDSSs into practice remains limited due to challenges of transparency, validation, clinical trust, and workflow compatibility. \cite{harris2023large}.

\subsection{Adaptation and Training Strategies for LLMs}

Adapting LLMs to healthcare requires overcoming challenges of data scarcity, privacy constraints, and compute limitations. Full fine-tuning is often infeasible in clinical environments, prompting interest in parameter-efficient methods. Low-rank adaptation (LoRA) \cite{hu2022lora}, prompt tuning \cite{lester2021power}, and prefix tuning \cite{li2021prefix} enable targeted customization with reduced computational overhead. Instruction tuning \cite{wei2021finetuned, taori2023stanford} further aligns models to human-authored prompts, enhancing few-shot and zero-shot capabilities critical in data-sparse domains.

RAG \cite{lewis2020retrieval} adds an external memory component by conditioning generation model outputs on retrieved documents, improving factual grounding and traceability. Reinforcement learning from human feedback (RLHF) \cite{ouyang2022training} has been explored to align LLM model behavior with human values and clinical norms, though its application to structured medical decision-making remains early-stage. Our framework extends this line by integrating RAG with structured-unstructured data fusion to produce interpretable and personalized recommendations.

\subsection{Architectural Variants of Language Models}

Rapid advancements in LLM architecture have introduced diverse design trade-offs that are increasingly relevant to clinical use. Standard dense models such as GPT-3 and LLaMA \cite{touvron2023llama} offer strong performance but are costly in terms of inference and memory. Small language models (SLMs), including LLaMA 3B and T5-small \cite{raffel2020t5}, provide computational efficiency with only modest degradation in accuracy when aligned to domain-specific tasks—making them attractive for real-time clinical support.

Long-context models such as DeepSeek-R1 and OpenAI GPT-4 with 128K+ token capacity allow document-level reasoning and longitudinal history modeling, which are essential for tasks like summarizing hospital stays or cross-episode treatment planning \cite{jiang2023mistral}. Mixture-of-Experts (MoE) architectures such as Mixtral \cite{beeching2023mixtral} and DeepSeek-MoE dynamically route inputs through subsets of the model, balancing scalability with compute efficiency. These designs are promising for clinical workloads that vary dramatically in complexity.

State Space Models (SSMs) like Mamba \cite{gu2023mamba} show potential in long-sequence, low-latency scenarios and may serve as efficient alternatives to transformer-based models in time-sensitive clinical environments. RAG-based models like RETRO \cite{borgeaud2022retro}, which combine dense retrieval with language generation, introduce grounding mechanisms that support factual correctness and traceability—attributes that are vital for clinical trust and safety but are still underexplored in the healthcare context. Our proposed architecture draws inspiration from these innovations, adopting a RAG pipeline over structured and unstructured clinical inputs, while remaining compatible with parameter-efficient adaptation strategies and low-latency deployment environments.

While substantial progress has been made in adapting LLMs for biomedical NLP, existing models and architectures remain underutilized for real-time clinical decision support tasks, especially prescribing. Moreover, architectural trade-offs, retrieval strategies, and adaptation methods remain fragmented and poorly benchmarked in safety-critical healthcare environments. Our work aims to bridge this gap by proposing a practical, modular, and explainable framework that leverages retrieval-augmented language modeling over heterogeneous EHR data to assist prescribers with contextualized treatment recommendations.

Our work emphasizes integration with real or synthetic EHR data, the fusion of structured and unstructured clinical signals, and the provision of transparent, precedent-based reasoning in support of safe prescribing. We build upon the core insight that patient similarity—measured across clinical, demographic, and historical features—is a valuable basis for generating recommendations in a manner consistent with human expert practices.

\section{Conclusion}
This work presents a RAG-LLM framework for clinically grounded prescribing support, leveraging historical emergency department records to generate patient-specific treatment recommendations. By integrating structured and unstructured patient data with precedent-driven retrieval, our system provides context-sensitive outputs aligned with empirically observed clinical behavior. Evaluation on a dataset over 68,000 emergency department encounters demonstrate that RAG-LLMs perform competitively with traditional machine learning models in predicting treatment outcomes, while offering superior interpretability and clinical consistency through grounding in analogous cases. Notably, the model achieved strong alignment with observed prescriptions and generalized effectively across vulnerable subgroups, including uninsured and undomiciled patients. The proposed framework contributes both a scalable technical architecture and a clinically motivated evaluation protocol that together address limitations of existing black-box LLMs in healthcare. By assessing predictive accuracy retrieval quality and relaxed clinical agreement, we provide a multidimensional evaluation aligned with the realities of medical decision-making. Overall, this work demonstrates the feasibility and value of integrating LLM-powered decision support into prescribing workflows. By prioritizing explainability, transparency, and clinician oversight, RAG systems can serve as a foundation for more equitable, more personalized, and safet treatment delivery.

\bibliographystyle{ACM-Reference-Format}
\bibliography{sample-base}

\end{document}